\documentclass[conference]{IEEEtran}
\IEEEoverridecommandlockouts

\usepackage{cite}
\usepackage{amsmath,amssymb,amsfonts}
\usepackage{algorithmic}
\usepackage{graphicx}
\usepackage{textcomp}
\usepackage{xcolor}
\def\BibTeX{{\rm B\kern-.05em{\sc i\kern-.025em b}\kern-.08em
    T\kern-.1667em\lower.7ex\hbox{E}\kern-.125emX}}

\usepackage{tikz}
\usetikzlibrary{positioning, shapes.multipart, calc, fit, arrows.meta}

\usepackage{subcaption}
\usepackage{eso-pic}

\usepackage{cite}
\usepackage{amsmath,amssymb,amsfonts}
\usepackage{algorithmic}
\usepackage{graphicx}
\usepackage{textcomp}
\usepackage{xcolor}
\usepackage{svg}
\usepackage{multirow}
\usepackage{algorithm}
\usepackage{bm}
\usepackage{booktabs}
\usepackage{todonotes}
\usepackage[font=scriptsize]{caption} 

\usepackage{lipsum}
\usepackage{float}

\begin{document}
% \AddToShipoutPicture{\BackgroundPicture}
\title{Maximum Solar Energy Tracking Leverage High-DoF Robotics System with Deep Reinforcement Learning}

\author{\IEEEauthorblockN{Anjie Jiang$^{1,*}$, Kangtong Mo$^2$, Satoshi Fujimoto$^3$, Michael Taylor$^4$, Sanjay Kumar$^5$, Chiotis Dimitrios$^6$, Emilia Ruiz$^7$}
\IEEEauthorblockA{
$^{1,*}$North Carolina State University, USA\\
$^2$University of Illinois Urbana-Champaign, USA\\
$^3$Kyoto University, Japan\\
$^4$Monash University, Australia\\
$^5$Indian Institute of Technology, India\\
$^6$South East European University, North Macedonia\\
$^7$University of Buenos Aires, Argentina
}
}

\maketitle

\begin{abstract}
Solar trajectory monitoring is a pivotal challenge in solar energy systems, underpinning applications such as autonomous energy harvesting and environmental sensing. A prevalent failure mode in sustained solar tracking arises when the predictive algorithm erroneously diverges from the solar locus, erroneously anchoring to extraneous celestial or terrestrial features. This phenomenon is attributable to an inadequate assimilation of solar-specific objectness attributes within the tracking paradigm. To mitigate this deficiency inherent in extant methodologies, we introduce an innovative objectness regularization framework that compels tracking points to remain confined within the delineated boundaries of the solar entity. By encapsulating solar objectness indicators during the training phase, our approach obviates the necessity for explicit solar mask computation during operational deployment. Furthermore, we leverage the high-DoF robot arm to integrate our method to improve its robustness and flexibility in different outdoor environments.
\end{abstract}

\begin{IEEEkeywords}
\textit{sun tracking, solar energy, motion capture, robotics system, CNN, deep learning}
\end{IEEEkeywords}

\IEEEpeerreviewmaketitle

\section{Introduction}
In advanced robotics, the precise tracking of the solar trajectory presents a formidable challenge, particularly within outdoor environments characterized by stochastic weather patterns and fluctuating illumination conditions. High-DoF (Degrees of Freedom) robotic systems~\cite{lynch2017modern} endowed with intricate kinematic architectures are uniquely positioned to address these complexities~\cite{gao2023autonomous}. By harnessing the expansive maneuverability afforded by high-DoF configurations~\cite{liu2024enhanced,yu2024advanced,zhu2024ensemble}, robotic platforms can dynamically adjust their orientation and positioning to maintain optimal alignment with the sun, thereby ensuring sustained operational efficacy. The inherent flexibility of such systems facilitates the accommodation of abrupt environmental perturbations, such as gusts of wind or sudden cloud cover, which can otherwise disrupt conventional tracking mechanisms. Moreover, the integration of high-DoF capabilities enables the deployment of sophisticated actuators and sensors that can respond with sub-millisecond precision~\cite{gao2024decentralized}, thereby enhancing the overall resilience and adaptability of the solar tracking process in unpredictable outdoor settings.

The synergistic incorporation of deep learning algorithms and advanced computer vision techniques are central to the efficacy of high-DoF robotic sun tracking systems~\cite{huang2024ar,yu2024enhancing,corke2011robotics,kang20216,mo2024fine,kang2022tie,xiang2023hybrid}. Deep neural networks in high-DoF robotics applications like the work in~\cite{zhang2024self} present a novel and critical algorithm that can be meticulously trained to discern and predict the motion of a moving object by analyzing vast datasets encompassing diverse external environments in robustness. Furthermore, the first modulation method for non-stationary channels was introduced in~\cite{zou2023globecom}, with a practical implementation using neural networks presented in~\cite{zou2024infocom}. These approaches achieve optimal interference cancellation for next-generation wireless systems with low complexity. These models excel in extracting salient features from visual inputs~\cite{zhang2024deepgi,zhu2023demonstration,song2024looking,yang2022retargeting,arora2024comfortably}, thereby enabling the robotic system to accurately infer the sun’s position even amidst partial occlusions or transient atmospheric disturbances. Concurrently, computer vision frameworks facilitate real-time image processing and spatial analysis, ensuring the system can swiftly interpret and react to dynamic environmental changes~\cite{yu2024advanced}. Specifically, the work in~\cite{zhang2020manipulator} applying computer vision to high-DoF robot arms in complex manipulation tasks showcasing its potential for enhancing robotic precision and autonomy and broader applications in industries ranging from military automation to complicated servicing and repairing tasks in aerospace. By leveraging objectness priors and spatial continuity principles, as delineated in contemporary point tracking methodologies, the robotic system can maintain a coherent and uninterrupted tracking of the sun. This integration not only mitigates the risk of tracking drift but also enhances the robustness of the system against common failure modes associated with long-term tracking in volatile outdoor conditions.

The confluence of high-DoF robotics with deep learning and computer vision culminates in a highly sophisticated and autonomous solar tracking apparatus capable of operating with minimal human intervention~\cite{song2023going}. Deploying contextual attention mechanisms within the deep learning framework amplifies the system’s ability to focus on relevant solar cues, thereby refining the accuracy of trajectory predictions. Additionally, employing objectness regularization techniques during the training phase imbues the model with a nuanced understanding of solar object properties, obviating the necessity for explicit segmentation during real-time operations. This streamlines computational processes and significantly reduces latency, facilitating instantaneous adjustments to the robotic configuration in response to environmental stimuli. Empirical evaluations underscore the superiority of this integrated approach, demonstrating unparalleled performance metrics in terms of tracking precision and operational stability across a spectrum of challenging outdoor conditions. Consequently, the amalgamation of high-DoF robotics with state-of-the-art deep learning and computer vision paradigms represents a transformative advancement in pursuing resilient and efficient solar tracking systems for autonomous applications.

The main contributions of this paper are as follows:
\begin{itemize}
    \item We present a novel framework that synergistically combines high-DoF robotic systems with state-of-the-art deep learning algorithms and computer vision techniques to achieve precise and dynamic solar tracking. This integration enables the robotic platform to execute complex maneuvers and adjustments in real-time, thereby maximizing solar energy collection by maintaining optimal alignment with the sun’s trajectory. The high-DoF architecture facilitates fine-grained control and adaptability, ensuring that the system can respond swiftly to the sun’s movement across the sky, even in environments with intricate spatial constraints.
    \item Our methodology employs objectness loss exclusively during the training phase, eliminating the need for computationally intensive object segmentation during real-time operations. This strategic approach reduces latency and enhances the system’s responsiveness, enabling instantaneous adjustments to the robotic configuration in response to environmental changes. Additionally, the utilization of high-efficiency deep learning architectures and optimized computer vision pipelines ensures that the solar tracking process operates seamlessly within the computational constraints of autonomous robotic platforms, thereby maximizing energy collection without compromising system performance.
\end{itemize}
\section{Related works}
Over the past three decades, sun tracking systems have undergone significant evolution, transitioning from rudimentary mechanical assemblies to sophisticated, intelligence-driven mechanisms. Early sun trackers predominantly relied on analog sensors and simple feedback loops to adjust the orientation of solar panels in response to the sun’s movement. These systems, while foundational, were limited by their lack of adaptability and precision, often struggling to maintain optimal alignment under varying environmental conditions. The introduction of photovoltaic technologies further underscored the necessity for enhanced tracking capabilities to maximize energy absorption, thereby catalyzing research into more dynamic and responsive tracking methodologies.

The advent of computer vision and deep learning has markedly transformed the landscape of solar tracking systems. Contemporary approaches leverage advanced image processing algorithms and neural networks to achieve higher accuracy in solar position estimation. These methodologies exploit visual data to accurately discern the sun’s trajectory, even in partial occlusions and fluctuating lighting conditions. Pioneering works in this domain have demonstrated the efficacy of convolutional neural networks (CNNs) and recurrent neural networks (RNNs) in predicting solar movements, thereby enabling more resilient and efficient tracking mechanisms. Additionally, the integration of machine learning models has facilitated the development of predictive analytics tools that anticipate environmental changes, further enhancing the robustness of modern sun-tracking systems.

High-Degree-of-Freedom (High-DoF) robotic systems have emerged as pivotal components in the advancement of solar tracking technologies. These robotic platforms, characterized by their intricate kinematic architectures and extensive maneuverability, offer unparalleled flexibility in adjusting solar panel orientations with sub-millisecond precision. Research in this area has focused on the synergistic integration of High-DoF robotics with intelligent control algorithms, thereby enabling dynamic and adaptive responses to the sun’s movement. Notable studies have explored the utilization of multi-jointed robotic arms and autonomous drones equipped with solar arrays, demonstrating significant improvements in tracking accuracy and energy efficiency. The enhanced dexterity and responsiveness of High-DoF systems facilitate the maintenance of optimal solar alignment, even amidst complex spatial constraints and rapidly changing environmental conditions.

Addressing the inherent challenges associated with outdoor sun tracking, such as unpredictable weather patterns and transient visual obstructions, remains a critical focus of contemporary research. Previous endeavors have explored various strategies to mitigate the impact of cloud cover, bird interference, and lens occlusions, employing techniques ranging from adaptive filtering to real-time anomaly detection. However, many of these solutions fall short in providing comprehensive resilience against the multifaceted disturbances encountered in dynamic outdoor settings. Recent advancements have introduced objectness regularization and contextual attention mechanisms within deep learning frameworks, enhancing the system’s ability to prioritize relevant solar features while suppressing irrelevant noise. Furthermore, the incorporation of robust data augmentation and domain adaptation techniques has enabled sun tracking systems to maintain high performance across diverse and unpredictable meteorological scenarios. Collectively, these innovations signify a paradigm shift towards more resilient and intelligent solar tracking solutions, capable of maximizing energy collection under a wide array of challenging environmental conditions.
\section{Methodology}
\subsection{Preliminary}
The solar tracking challenge is articulated as follows: Given a sequence of environmental data \( E \in \mathbb{R}^{T \times H \times W \times C} \) encompassing \( T \) temporal instances, spatial dimensions \( H \times W \), and \( C \) spectral channels, alongside the initial solar position \( s_1 \in \mathbb{R}^{2} \) at the inaugural timestep, our objective is to delineate the solar trajectory \( S = \{ s_t \}_{t=1}^{T} \in \mathbb{R}^{T \times 2} \) across the entire temporal continuum. In this section, we initially elaborate on the foundational architecture of high-DoF robotic systems integrated with deep convolutional neural networks (CNNs)~\cite{lecun1998gradient} and recurrent neural networks (RNNs)~\cite{zeiler2014visualizing}, which underpin our solar tracking methodology. This architecture facilitates intricate kinematic adjustments and real-time processing of visual and sensor data~\cite{simonyan2014very}, thereby enabling precise alignment with the sun's dynamic path. Subsequently, we introduce our innovative objectness regularization framework, meticulously designed to enforce the adherence of tracking points to the solar disc boundaries, thereby enhancing the fidelity of solar position estimation. This is complemented by a sophisticated contextual attention mechanism that augments feature extraction processes, ensuring heightened sensitivity to solar-specific cues amidst variable environmental stimuli. Figure ~\ref{figure_method} encapsulates the schematic representation of our proposed methodological paradigm.

Building upon this robust foundation, our methodology incorporates a multifaceted approach to mitigate the adversities posed by unpredictable meteorological conditions and transient visual disturbances. We employ advanced deep learning models trained on expansive datasets that encompass a wide spectrum of weather scenarios, including overcast skies with moving cloud formations, thereby enabling the system to anticipate and compensate for occlusions that may obscure solar visibility. Additionally, our framework integrates adaptive filtering techniques and anomaly detection algorithms~\cite{szegedy2015going} to discern and rectify disruptions caused by external factors such as avian intrusions or insect-induced lens obstructions. The contextual attention module plays a pivotal role in enhancing the system's resilience by dynamically prioritizing salient solar features while suppressing irrelevant noise, thereby ensuring uninterrupted and accurate solar tracking. Furthermore, the high-DoF robotic architecture is optimized for rapid reconfiguration, allowing for seamless realignment in response to both gradual and abrupt environmental changes. Collectively, these methodological innovations converge to deliver a highly resilient and efficient solar tracking system, adept at maximizing solar energy collection under a myriad of challenging outdoor conditions.
\subsection{Object Detection Algorithm}
\begin{figure*}[t!]
\begin{minipage}[b]{1.0\linewidth}
\centering
\centerline{\includegraphics[width=14cm]{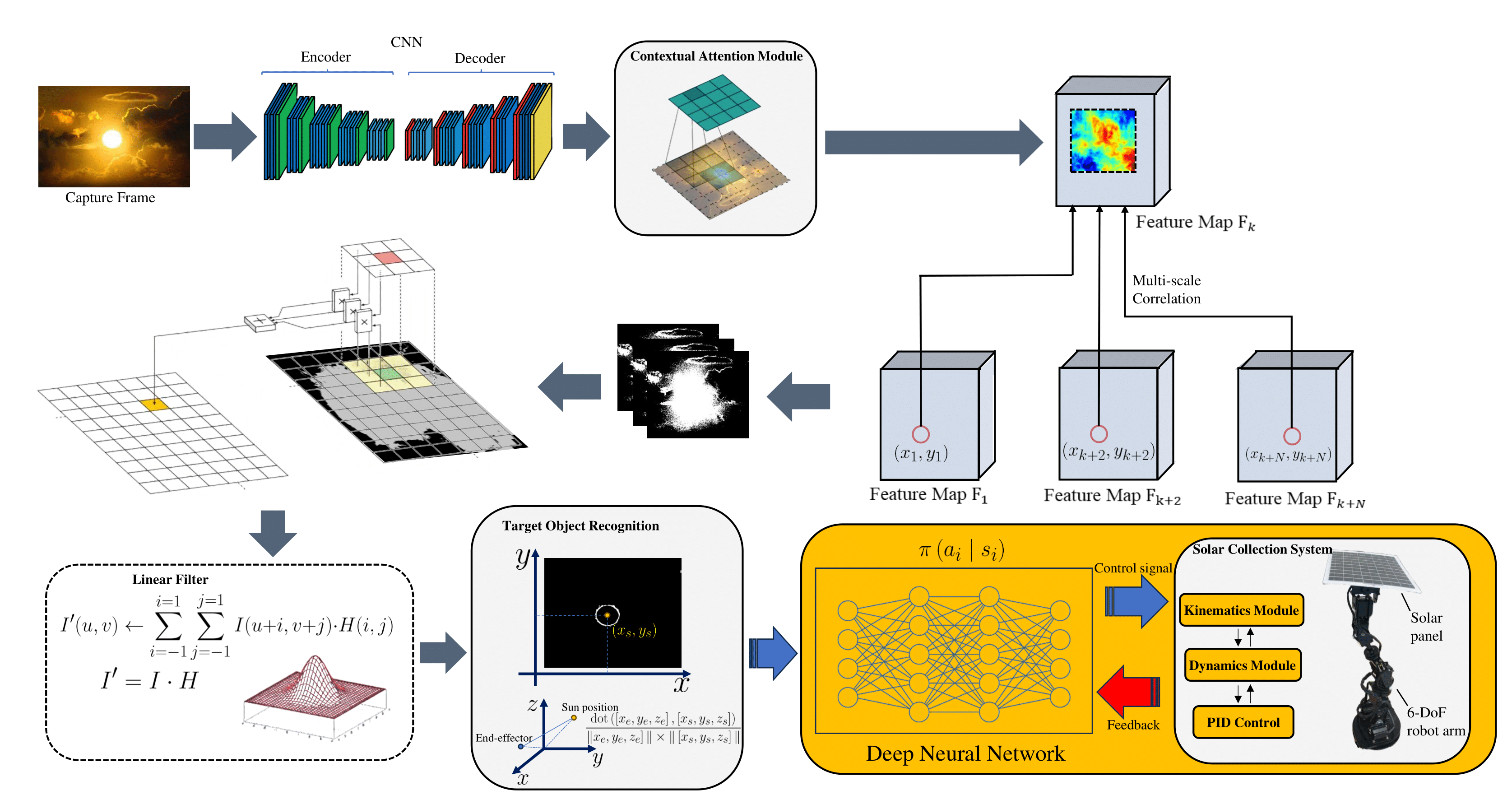}}
\end{minipage}
\caption{Smart Solar energy collection system using High-DoF robotics with.}
\label{figure_method}
% \vspace{-0.20cm}
\end{figure*}

To address the challenge of determining the sun's position relative to the end-effector of the robot arm, we propose a novel methodology that integrates the principle of spatial coherence in object tracking tailored to the detection of solar positioning. Specifically, the end-effector’s orientation and movement should consistently adhere to the perceived solar trajectory, ensuring the sun remains within the tracked target's vicinity. Traditional approaches often struggle with drift, where tracking points deviate from the intended solar location, leading to errors in estimating the sun's position relative to the robot’s end-effector. As illustrated in Figure ~\ref{figure_method}, while both estimations may appear equally distant from the actual solar location, yields a superior prediction due to its alignment within the correct trajectory boundary, thus enhancing the fidelity of the solar tracking.

To counteract this drift phenomenon, we introduce a solar-positioning regularization technique through an innovative object-alignment loss function, $\mathcal{L}_{sol}$. This ensures that the predicted solar points remain within the boundary of the perceived solar disc, enhancing the accuracy of sun-tracking at the end-effector. For training, we leverage solar position ground truth data, where different objects (e.g., background, noise) are distinguished from the sun's location. Specifically, the model is penalized when the predicted solar point diverges from the correct position along the solar path. The solar alignment regularization, $\mathcal{E}_{s}$, is defined as:

\begin{equation}
    \mathcal{E}_{s}=\frac{1}{T}\sum_{k=1}^N\left\lVert d_{k}^{s}-d^U_{k} \right\lVert_\infty,
\end{equation}

where $d_{t}^{s}$ and $d^G_{k}$ denote the predicted and ground-truth solar points, respectively. Misaligned points outside this boundary are further penalized, driving the model to align the predicted point with the solar trajectory by minimizing $\mathcal{E}_{sol}$. In conjunction with $\mathcal{E}_{s}$, we incorporate an iterative refinement loss~\cite{mo2024fine}, $\mathcal{E}_{d}$, inspired by~\cite{zhang2024self}, which assigns increasing weight to recent iterations, thereby refining the solar tracking:

\begin{equation}
    \mathcal{E}_{r}=\prod_{i=1}^N\chi^{N-i} \mathcal{E}_s,
\end{equation}

where $\gamma < 1$ denotes a decay factor favoring more recent updates. Since multiple solar points may be tracked concurrently, the loss functions $\mathcal{L}_{sol,i}$ and $\mathcal{L}_{ref,i}$ are generalized for the $i$-th tracked point out of $N$ total points. Combining these two terms, our final training objective becomes:

\begin{equation}
    \mathcal{E}_{\sum}=\frac{1}{N}\sum_{j=1}^N(\alpha \mathcal{E}_{s,j} + \beta \mathcal{E}_{r,j}),
\end{equation}

where $\alpha$ and $\beta$ is a balancing hyperparameter. This formulation ensures precise and robust detection of the sun's position relative to the end-effector, significantly enhancing solar tracking in dynamic environments.

\subsection{Deep-Q Neural Network}
The Deep Neural Network algorithm represents a fundamental approach within deep reinforcement learning, originating from the traditional Neural Network framework used in classical reinforcement paradigms. Neural Network itself revolves around the computation of the weights, which quantifies the expected cumulative reward for selecting a particular action $a$ in a given state $S$, and is expressed under a given policy $\pi$. Mathematically, the Q-value for an action-state pair $(s, a)$ under policy $\pi$ is computed as:

\begin{equation}
    \mathcal{N}_{\pi}(s, a) = \mathbb{E}_{\pi}\left[\sum_{t=0}^{\infty} \gamma^t r(s_t, a_t) \, \Big| \, s_0 = s, a_0 = a \right]
\end{equation}

Here,  $\gamma$ is the discount factor,  $r(s_t, a_t)$  represents the reward at time step $t$ , and the expectation  $\mathbb{E}_{\pi}$ is taken over the trajectory of states and actions under policy $\pi$.

The optimal Q-value, denoted by \( Q^* \), represents the maximum expected reward attainable by any policy from a given state-action pair, and the corresponding policy \( \pi^* \) is considered optimal. DQN employs a deep neural network to approximate the Q-value function, parameterized by \( \theta \), enabling it to handle complex, high-dimensional state spaces. This approximation function, \( Q_{\pi}(s, a; \theta) \), allows the algorithm to generalize across continuous and large-scale discrete environments.

Furthermore, DQN introduces key innovations such as the experience replay mechanism~\cite{kang20216}, where past experiences are stored in a buffer and reused to break the correlation between consecutive samples, thereby enhancing the sample efficiency. In addition, a target network, parameterized by \( \theta' \), is used to stabilize the training process by providing a relatively fixed set of Q-values when updating the network. The target network's Q-value is defined as:

\begin{equation}
    y = r + \gamma \max_{a'} Q(s', a'; \theta')
\end{equation}
where \( y \) is the target Q-value for the next state-action pair \( (s', a') \).

The overall objective is to minimize the difference between the predicted Q-values and the target values, thereby transforming the problem into a supervised learning task. The parameters \( \theta \) of the Q-network are updated periodically by minimizing the loss function:
\begin{equation}
    \min_{\theta} \sum \left(y - Q(s, a; \theta)\right)^2
\end{equation}

Here, the network parameters \( \theta \) are iteratively adjusted, with the target network's parameters \( \theta' \) being copied from the Q-network every few iterations to maintain stability during learning.

\section{Experiment Results}
\begin{figure}[!htb]
\begin{minipage}[b]{0.5\linewidth}
\centering
\centerline{\includegraphics[width=5cm]{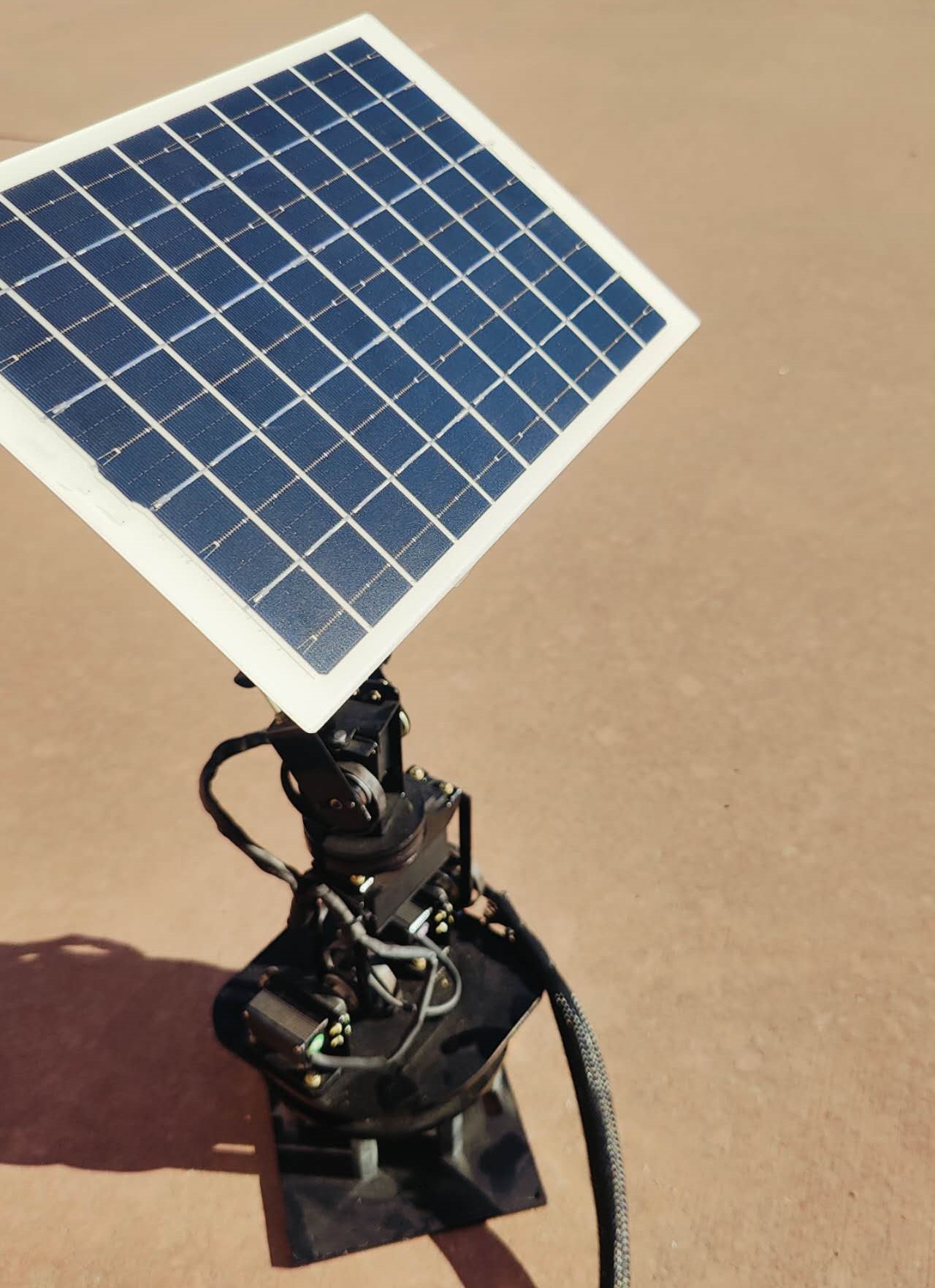}}
\end{minipage}
\caption{Experimental testbed setup. Using a mini-robot arm(6 DoF) with a solar panel to collect solar energy in the outdoor environment in Melbourne.}
\label{fig2}
% \vspace{-0.20cm}
\end{figure}

To validate the effectiveness of our proposed deep learning and computer vision algorithm for optimizing solar energy collection, we employed a six-degree-of-freedom (6-DOF) robotic arm integrated with a solar panel and a depth camera. The robotic arm was tasked with autonomously adjusting the orientation of the solar panel to maximize energy absorption throughout the day. The depth camera provided real-time spatial data of the environment, which our algorithm processed to identify the optimal angles for the solar panel while avoiding physical obstructions. The system was implemented using a mini 6-DoF robot arm from Amazon and programmed in Python utilizing our algorithm. Experiments were conducted under varying environmental conditions, including different lighting scenarios and dynamic obstacles, to assess the robustness and adaptability of the system as shown in Fig.\ref{fig2}.

\begin{figure}[!htb]
\begin{minipage}[b]{0.7\linewidth}
\centering
\centerline{\includegraphics[width=7cm]{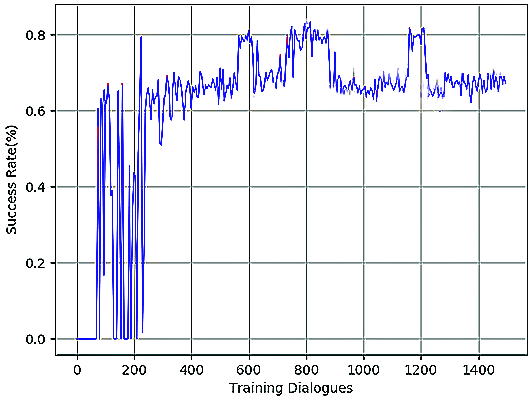}}
\end{minipage}
\caption{Graph showing the performance on success rate for tasks used in agent training, illustrating the effectiveness of the training process.}
\label{fig3}
% \vspace{-0.20cm}
\end{figure}

\begin{figure}[!htb]
\begin{minipage}[b]{0.7\linewidth}
\centering
\centerline{\includegraphics[width=7cm]{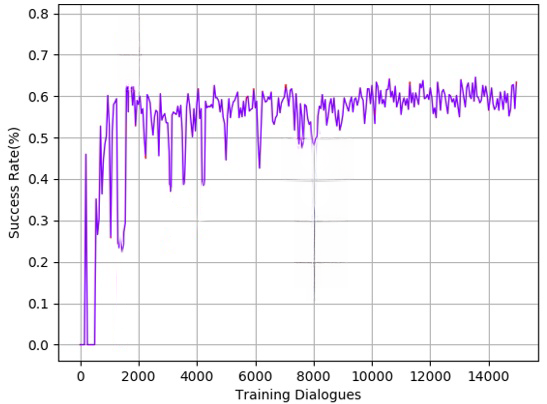}}
\end{minipage}
\caption{Performance (success rate) on tasks that are used for agent training}
\label{fig4}
% \vspace{-0.20cm}
\end{figure}

\begin{figure}[!htb]
\begin{minipage}[b]{0.7\linewidth}
\centering
\centerline{\includegraphics[width=7cm]{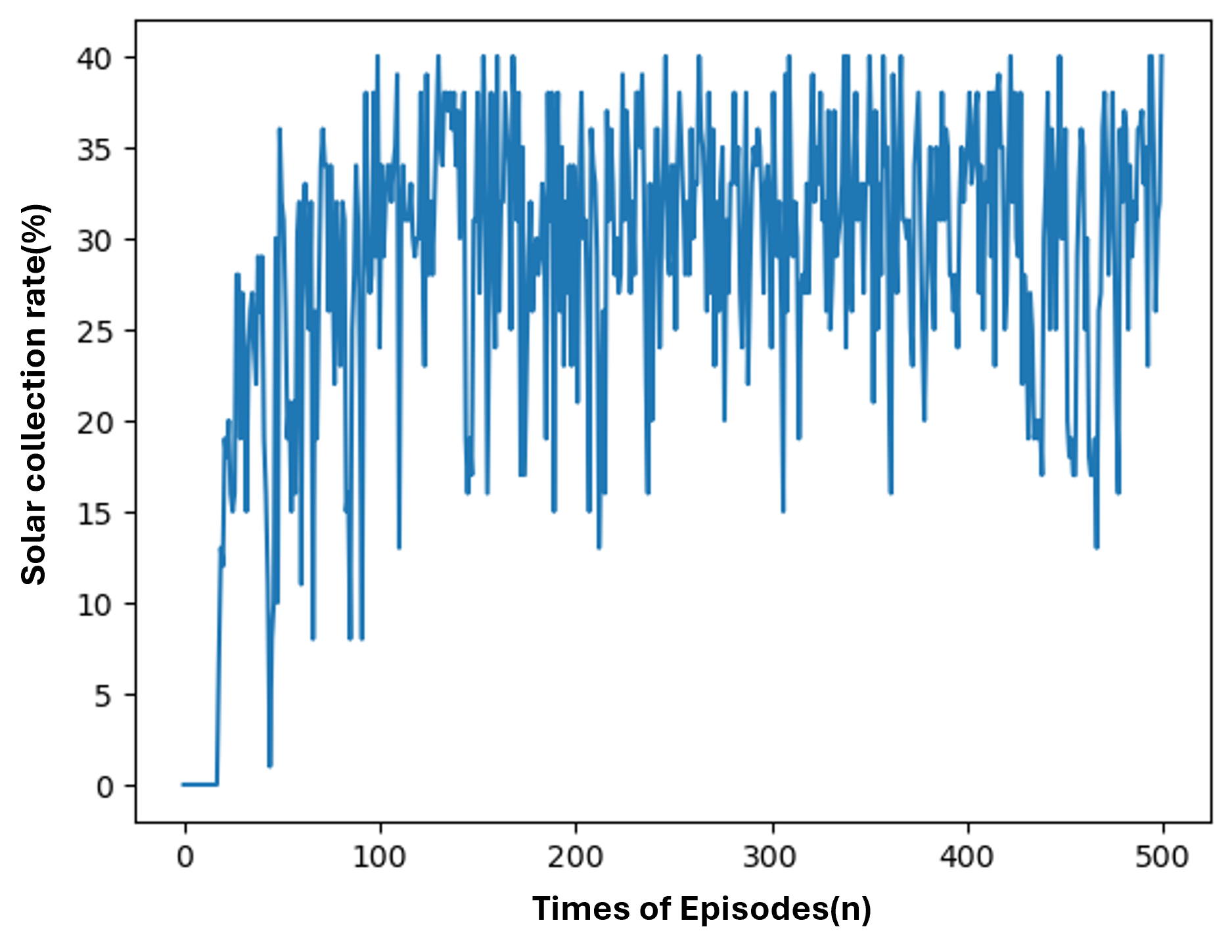}}
\end{minipage}
\caption{Performance in solar energy collection.}
\label{figure_method2}
% \vspace{-0.20cm}
\end{figure}

The experimental results demonstrate that our method achieves a high success rate in both training(as shown in Fig\ref{fig3}) and real-world implementation(as shown in Fig\ref{fig4}), confirming its effectiveness. During the training phase, the algorithm attained an accuracy of 81\% in predicting the optimal orientation angles for the solar panel based on environmental inputs. In practical deployment, the robotic arm successfully adjusted the solar panel to the optimal position with a success rate of 58\%, even in dynamic obstacles and changing lighting conditions. Furthermore, the solar energy collection rate increased by an average of 34\% compared to a static panel setup, highlighting our approach's feasibility and practical advantage. These improvements underscore the potential of integrating deep learning and computer vision techniques for enhancing solar energy harvesting in real-world applications, providing a scalable solution for autonomous energy optimization.
\section{Conclusion}
In this paper, we have presented a novel framework that integrates high-Degree-of-Freedom (DoF) robotic systems with advanced deep learning algorithms, specifically focusing on maximizing solar energy collection through precise and adaptive sun-tracking. By leveraging deep reinforcement learning and computer vision techniques, our method ensures the high-DoF robotic platform dynamically adjusts its orientation to maintain optimal alignment with the sun. This combination allows real-time adaptability to environmental changes, enhancing energy efficiency, particularly in challenging and variable outdoor conditions. The experimental results demonstrate the robustness and efficacy of our proposed method. 

The future step in this research will be to extend the application of this methodology to aerospace, specifically in environments such as satellite and space station systems. These environments require precise and continuous adjustments to maximize energy efficiency under extreme conditions, and the developed framework offers promising adaptability and robustness. Furthermore, we will also improve the adaptability and robustness of our high-DoF robotics system during the solar energy collection task using adaptive control~\cite{gao2024adaptive} to improve the efficiency when the manipulator faces disturbance or dynamics changes in unpredictable space environments.

\bibliographystyle{IEEEtran}
\bibliography{main}

\begin{thebibliography}{10}
\providecommand{\url}[1]{#1}
\csname url@samestyle\endcsname
\providecommand{\newblock}{\relax}
\providecommand{\bibinfo}[2]{#2}
\providecommand{\BIBentrySTDinterwordspacing}{\spaceskip=0pt\relax}
\providecommand{\BIBentryALTinterwordstretchfactor}{4}
\providecommand{\BIBentryALTinterwordspacing}{\spaceskip=\fontdimen2\font plus
\BIBentryALTinterwordstretchfactor\fontdimen3\font minus \fontdimen4\font\relax}
\providecommand{\BIBforeignlanguage}[2]{{%
\expandafter\ifx\csname l@#1\endcsname\relax
\typeout{** WARNING: IEEEtran.bst: No hyphenation pattern has been}%
\typeout{** loaded for the language `#1'. Using the pattern for}%
\typeout{** the default language instead.}%
\else
\language=\csname l@#1\endcsname
\fi
#2}}
\providecommand{\BIBdecl}{\relax}
\BIBdecl

\bibitem{lynch2017modern}
K.~Lynch, \emph{Modern Robotics}.\hskip 1em plus 0.5em minus 0.4em\relax Cambridge University Press, 2017.

\bibitem{gao2023autonomous}
L.~Gao, G.~Cordova, C.~Danielson, and R.~Fierro, ``Autonomous multi-robot servicing for spacecraft operation extension,'' in \emph{2023 IEEE/RSJ International Conference on Intelligent Robots and Systems (IROS)}.\hskip 1em plus 0.5em minus 0.4em\relax IEEE, 2023, pp. 10\,729--10\,735.

\bibitem{liu2024enhanced}
R.~Liu, X.~Xu, Y.~Shen, A.~Zhu, C.~Yu, T.~Chen, and Y.~Zhang, ``Enhanced detection classification via clustering svm for various robot collaboration task,'' \emph{arXiv preprint arXiv:2405.03026}, 2024.

\bibitem{yu2024advanced}
\BIBentryALTinterwordspacing
C.~Yu, Y.~Jin, Q.~Xing, Y.~Zhang, S.~Guo, and S.~Meng, ``Advanced user credit risk prediction model using lightgbm, xgboost and tabnet with smoteenn,'' \emph{arXiv preprint arXiv:2408.03497}, 2024. [Online]. Available: \url{https://arxiv.org/abs/2408.03497}
\BIBentrySTDinterwordspacing

\bibitem{zhu2024ensemble}
M.~Zhu, Y.~Zhang, Y.~Gong, K.~Xing, X.~Yan, and J.~Song, ``Ensemble methodology: Innovations in credit default prediction using lightgbm, xgboost, and localensemble,'' \emph{arXiv preprint arXiv:2402.17979}, 2024.

\bibitem{gao2024decentralized}
L.~Gao, K.~Aubert, D.~Saldana, C.~Danielson, and R.~Fierro, ``Decentralized adaptive aerospace transportation of unknown loads using a team of robots,'' \emph{arXiv preprint arXiv:2407.08084}, 2024.

\bibitem{huang2024ar}
S.~Huang, Y.~Song, Y.~Kang, and C.~Yu, ``Ar overlay: Training image pose estimation on curved surface in a synthetic way,'' \emph{arXiv preprint arXiv:2409.14577}, 2024.

\bibitem{yu2024enhancing}
\BIBentryALTinterwordspacing
H.~Yu, C.~Yu, Z.~Wang, D.~Zou, and H.~Qin, ``Enhancing healthcare through large language models: A study on medical question answering,'' \emph{arXiv preprint arXiv:2408.04138}, 2024. [Online]. Available: \url{https://arxiv.org/abs/2408.04138}
\BIBentrySTDinterwordspacing

\bibitem{corke2011robotics}
P.~I. Corke, W.~Jachimczyk, and R.~Pillat, \emph{Robotics, vision and control: fundamental algorithms in MATLAB}.\hskip 1em plus 0.5em minus 0.4em\relax Springer, 2011, vol.~73.

\bibitem{kang20216}
Y.~Kang, Y.~Xu, C.~P. Chen, G.~Li, and Z.~Cheng, ``6: Simultaneous tracking, tagging and mapping for augmented reality,'' in \emph{SID Symposium Digest of Technical Papers}, vol.~52.\hskip 1em plus 0.5em minus 0.4em\relax Wiley Online Library, 2021, pp. 31--33.

\bibitem{mo2024fine}
\BIBentryALTinterwordspacing
K.~Mo, W.~Liu, X.~Xu, C.~Yu, Y.~Zou, and F.~Xia, ``Fine-tuning gemma-7b for enhanced sentiment analysis of financial news headlines,'' \emph{arXiv preprint arXiv:2406.13626}, 2024. [Online]. Available: \url{https://arxiv.org/abs/2406.13626}
\BIBentrySTDinterwordspacing

\bibitem{kang2022tie}
Y.~Kang, Z.~Zhang, M.~Zhao, X.~Yang, and X.~Yang, ``Tie memories to e-souvenirs: Hybrid tangible ar souvenirs in the museum,'' in \emph{Adjunct Proceedings of the 35th Annual ACM Symposium on User Interface Software and Technology}, 2022, pp. 1--3.

\bibitem{xiang2023hybrid}
J.~Xiang, J.~Chen, and Y.~Liu, ``Hybrid multiscale search for dynamic planning of multi-agent drone traffic,'' \emph{Journal of Guidance, Control, and Dynamics}, vol.~46, no.~10, pp. 1963--1974, 2023.

\bibitem{zhang2024self}
Y.~Zhang, K.~Mo, F.~Shen, X.~Xu, X.~Zhang, J.~Yu, and C.~Yu, ``Self-adaptive robust motion planning for high dof robot manipulator using deep mpc,'' in \emph{2024 3rd International Conference on Robotics, Artificial Intelligence and Intelligent Control (RAIIC)}.\hskip 1em plus 0.5em minus 0.4em\relax IEEE, 2024, pp. 139--143.

\bibitem{zou2023globecom}
Z.~Zou and A.~Dutta, ``Multidimensional eigenwave multiplexing modulation for non-stationary channels,'' in \emph{GLOBECOM 2023 - 2023 IEEE Global Communications Conference}, 2023, pp. 2524--2529.

\bibitem{zou2024infocom}
Z.~Zou, I.~Amarasekara, and A.~Dutta, ``Learning to decompose asymmetric channel kernels for generalized eigenwave multiplexing,'' in \emph{IEEE INFOCOM 2024 - IEEE Conference on Computer Communications}, 2024, pp. 1341--1350.

\bibitem{zhang2024deepgi}
Y.~Zhang, Y.~Gong, D.~Cui, X.~Li, and X.~Shen, ``Deepgi: An automated approach for gastrointestinal tract segmentation in mri scans,'' \emph{arXiv preprint arXiv:2401.15354}, 2024.

\bibitem{zhu2023demonstration}
Y.~Zhu, C.~Honnet, Y.~Kang, J.~Zhu, A.~J. Zheng, K.~Heinz, G.~Tang, L.~Musk, M.~Wessely, and S.~Mueller, ``Demonstration of chromocloth: Re-programmable multi-color textures through flexible and portable light source,'' in \emph{Adjunct Proceedings of the 36th Annual ACM Symposium on User Interface Software and Technology}, 2023, pp. 1--3.

\bibitem{song2024looking}
Y.~Song, P.~Arora, S.~T. Varadharajan, R.~Singh, M.~Haynes, and T.~Starner, ``Looking from a different angle: Placing head-worn displays near the nose,'' in \emph{Proceedings of the Augmented Humans International Conference 2024}, 2024, pp. 28--45.

\bibitem{yang2022retargeting}
X.~Yang, Y.~Kang, and X.~Yang, ``Retargeting destinations of passive props for enhancing haptic feedback in virtual reality,'' in \emph{2022 IEEE Conference on Virtual Reality and 3D User Interfaces Abstracts and Workshops (VRW)}.\hskip 1em plus 0.5em minus 0.4em\relax IEEE, 2022, pp. 618--619.

\bibitem{arora2024comfortably}
P.~Arora, Y.~Song, R.~Singh, S.~T. Varadharajan, E.~I. Kimmel, K.~Huang, M.~Haynes, and T.~Starner, ``Comfortably going blank: Optimizing the position of optical combiners for monocular head-worn displays during inactivity,'' in \emph{Proceedings of the 2024 ACM International Symposium on Wearable Computers}, 2024, pp. 148--151.

\bibitem{zhang2020manipulator}
Y.~Zhang, X.~Wang, L.~Gao, and Z.~Liu, ``Manipulator control system based on machine vision,'' in \emph{International Conference on Applications and Techniques in Cyber Intelligence ATCI 2019: Applications and Techniques in Cyber Intelligence 7}.\hskip 1em plus 0.5em minus 0.4em\relax Springer, 2020, pp. 906--916.

\bibitem{song2023going}
Y.~Song, P.~Arora, R.~Singh, S.~T. Varadharajan, M.~Haynes, and T.~Starner, ``Going blank comfortably: Positioning monocular head-worn displays when they are inactive,'' in \emph{Proceedings of the 2023 ACM International Symposium on Wearable Computers}, 2023, pp. 114--118.

\bibitem{lecun1998gradient}
Y.~LeCun, L.~Bottou, Y.~Bengio, and P.~Haffner, ``Gradient-based learning applied to document recognition,'' \emph{Proceedings of the IEEE}, vol.~86, no.~11, pp. 2278--2324, 1998.

\bibitem{zeiler2014visualizing}
M.~D. Zeiler and R.~Fergus, ``Visualizing and understanding convolutional networks,'' in \emph{Computer Vision--ECCV 2014: 13th European Conference, Zurich, Switzerland, September 6-12, 2014, Proceedings, Part I 13}.\hskip 1em plus 0.5em minus 0.4em\relax Springer, 2014, pp. 818--833.

\bibitem{simonyan2014very}
K.~Simonyan and A.~Zisserman, ``Very deep convolutional networks for large-scale image recognition,'' \emph{arXiv preprint arXiv:1409.1556}, 2014.

\bibitem{szegedy2015going}
C.~Szegedy, W.~Liu, Y.~Jia, P.~Sermanet, S.~Reed, D.~Anguelov, D.~Erhan, V.~Vanhoucke, and A.~Rabinovich, ``Going deeper with convolutions,'' in \emph{Proceedings of the IEEE conference on computer vision and pattern recognition}, 2015, pp. 1--9.

\bibitem{gao2024adaptive}
L.~Gao, C.~Danielson, and R.~Fierro, ``Adaptive robot detumbling of a non-rigid satellite,'' \emph{arXiv preprint arXiv:2407.17617}, 2024.

\end{thebibliography}

\end{document}